\documentclass[letterpaper]{article} 
\usepackage{aaai2026}  
\usepackage{times}  
\usepackage{helvet}  
\usepackage{courier}  
\usepackage[hyphens]{url}  
\usepackage{graphicx} 
\usepackage{natbib}  
\usepackage{caption} 
\usepackage{algorithm}
\usepackage{algorithmic}

\usepackage{amsmath}      
\usepackage{autobreak}
\usepackage{amssymb}      
\usepackage{bm}  

\usepackage{booktabs}    
\usepackage{multirow}    
\usepackage{array}       
\usepackage{caption}     
\usepackage{newfloat}
\usepackage{listings}
\DeclareCaptionStyle{ruled}{labelfont=normalfont,labelsep=colon,strut=off} 
\floatstyle{ruled}
\newfloat{listing}{tb}{lst}{}
\floatname{listing}{Listing}

\title{From Classification to Ranking: Enhancing LLM Reasoning 
Capabilities for MBTI Personality Detection}
\author{
    Yuan Cao\textsuperscript{\rm 1},
    Feixiang Liu\textsuperscript{\rm 2},
    Xinyue Wang\textsuperscript{\rm 1}, 
    Yihan Zhu\textsuperscript{\rm 3},
    Hui Xu\textsuperscript{\rm 1},
    Zheng Wang\textsuperscript{\rm 1},
    Qiang Qiu\textsuperscript{\rm 1}}
\affiliations{
    \textsuperscript{\rm 1}Institute of Computing Technology, Chinese Academy of Sciences\\
    \textsuperscript{\rm 2}University of Chinese Academy of Sciences\\
    \textsuperscript{\rm 3}Beihang University\\
    {caoyuan, wangxinyue, xuhui, wangzheng, qiuqiang}@ict.ac.cn, liufeixiang23s@mails.ucas.ac.cn, zhuyihan@buaa.edu.cn
    }
\usepackage{bibentry}

\begin{document}

\maketitle

\begin{abstract}
Personality detection aims to measure an individual's corresponding personality traits through their social media posts. The advancements in Large Language Models (LLMs) offer novel perspectives for personality detection tasks. Existing approaches enhance personality trait analysis by leveraging LLMs to extract semantic information from textual posts as prompts, followed by training classifiers for categorization. However, accurately classifying personality traits remains challenging due to the inherent complexity of human personality and subtle inter-trait distinctions. Moreover, prompt-based methods often exhibit excessive dependency on expert-crafted knowledge without autonomous pattern-learning capacity. To address these limitations, we view personality detection as a ranking task rather than a classification and propose a corresponding reinforcement learning training paradigm. First, we employ supervised fine-tuning (SFT) to establish personality trait ranking capabilities while enforcing standardized output formats, creating a robust initialization. Subsequently, we introduce Group Relative Policy Optimization (GRPO) with a specialized ranking-based reward function. Unlike verification tasks with definitive solutions, personality assessment involves subjective interpretations and blurred boundaries between trait categories. Our reward function explicitly addresses this challenge by training LLMs to learn optimal answer rankings. Comprehensive experiments have demonstrated that our method achieves state-of-the-art performance across multiple personality detection benchmarks.
\end{abstract}


\section{Introduction}
Language-based personality detection is crucial for AI applications such as mental health screening and conversational agents, utilizing established psychological frameworks like Myers-Briggs Type Indicator (MBTI)  (Myers 1962\nocite{myers1962myers};  Myers and McCaulley 1985\nocite{rushton2012program}).  As illustrated in Figure 1, MBTI characterizes personality along four interdependent dimensions: Extraversion (E) vs. Introversion (I), Sensing (S) vs. iNtuition (N), Thinking (T) vs. Feeling (F), and Judging (J) vs. Perceiving (P).  Crucially, these dimensions exhibit dynamic interactions governed by hierarchical cognitive functions that collectively shape behavioral expressions.  For instance, in MBTI theory, an INTJ’s dominant function is introverted intuition (Ni), while ENTJ’s dominant function is extraverted thinking (Te), leading to different cognitive profiles even if only one letter differs. Prevailing approaches (Yang et al. 2021b\nocite{yang2021learning}; Zhu et al. 2024\nocite{zhu2024data}) fundamentally limit their efficacy by treating MBTI prediction as four independent binary classification tasks ("E/I", "S/N", "T/F", "J/P"), resulting in two critical shortcomings: dimensional decoupling that ignores psychometrically established interactions, static labeling that fails to capture complete personality expressions which neglects the position on the
scale (Stajner and Yenikent 2021\nocite{stajner2021why}). While traditional models struggle with this interdependence, the recent success of Large Language Models (LLMs) offers new opportunities to approach personality modeling in a more holistic and reasoning-driven manner.

Currently, LLMs exhibit remarkable reasoning capabilities across diverse NLP tasks (Brown et al. 2020\nocite{brown2020language}; Wei et al. 2022\nocite{wei2022chain}), but their application to personality assessment reveals fundamental limitations. Prompt-based methods such as PsyCoT (Yang et al. 2023b\nocite{yang2023psycot}) mimic the way individuals complete psychological questionnaires through multi-turn dialogues.  However, the limitations of prompt-based approaches are their over-reliance on expert-defined prompts and large-scale parameter models (Le et al. 2024\nocite{2024Adaptive}; Islam et al. 2025\nocite{islam2025promptbasedllmspositionbiasaware}), which makes them impractical for real-time applications due to high inference latency and computational cost. This gap necessitates novel frameworks capable of autonomously learning discriminative patterns from behavioral data.

\begin{figure}[t]
\centering
\includegraphics[width=0.9\columnwidth]{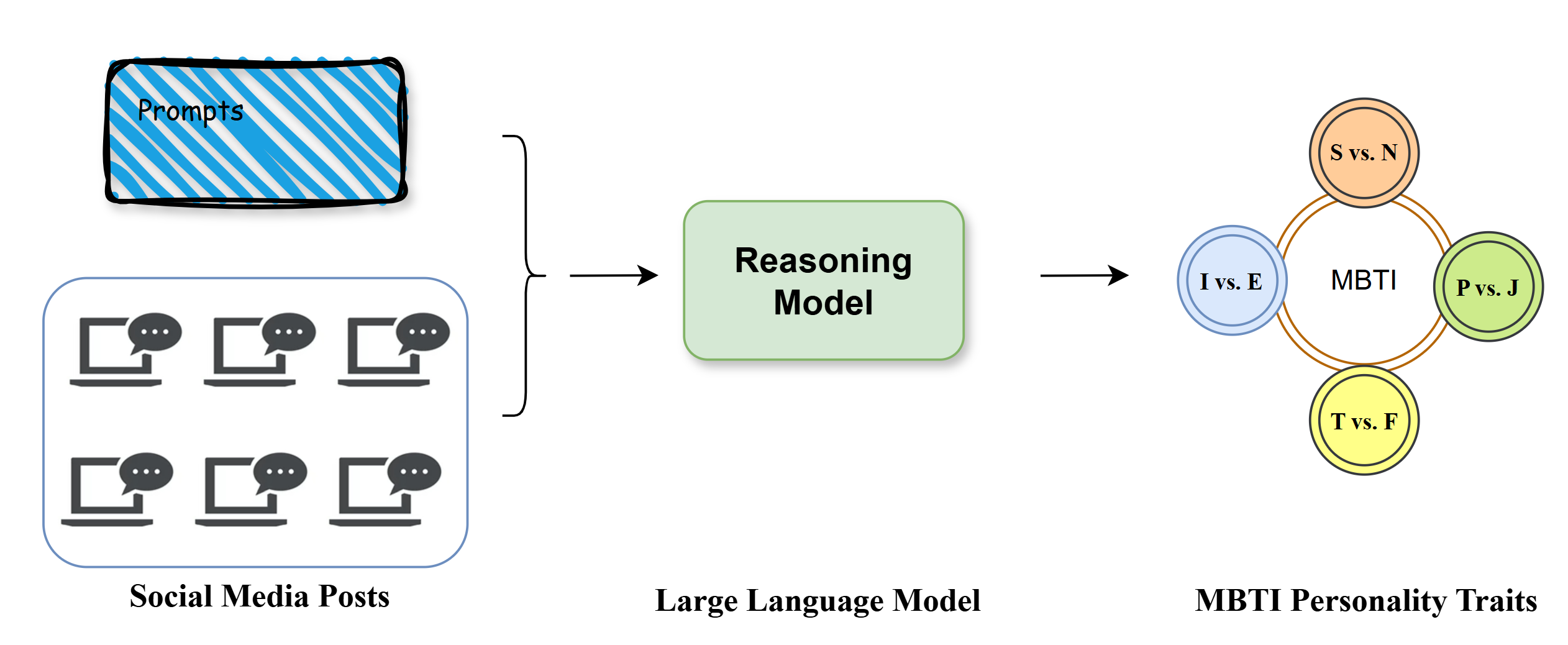}
\caption{A framework for LLM-based MBTI personality detection. The framework leverages a single user's social media posts as input to corresponding personality traits along four binary dimensions of Introversion vs. Extroversion, Sensing vs. iNtuition, Think vs. Feeling, and Perception vs. Judging.}
\label{fig1}
\end{figure}

To resolve these limitations, we pioneered a paradigm shift from categorical classification to preference ranking, a transformation aligned with psychometric realities where personality manifests itself through relative trait preferences rather than absolute categorical assignments (Heston, Thomas F and Gillette, Justin 2008\nocite{HULLERMEIER20081897}; Hüllermeier et al. 2025\nocite{Heston2025.03.14.25323987}).  This ranking-centric approach directly addresses the core challenge of modeling dimensional interdependencies by conceptualizing personality assessment as a continuous spectrum rather than discrete categories.  

Implementing this paradigm shift, we introduce PerDet-R1, a two-stage framework that eliminates categorical fallacies through ranking-based modeling.  In the first stage, we perform supervised fine-tuning (SFT) by distilling comparative reasoning traces from advanced LLMs Qwen-plus to generate ground-truth-aligned MBTI preference rankings from social media data, training a 7B-parameter LLM to output ordered type lists with explicit awareness of dimensional combinations and interactions.  The second stage implements Group Relative Policy Optimization (GRPO), initializing the policy with our stage-one model and optimizing ranking positions via a Normalized Discounted Cumulative Gain (NDCG) reward function, thereby amplifying trait-cohort interactions through position-sensitive reinforcement while eliminating expert prompt dependency.

In summary, our core contributions are as follows:

\begin{itemize}
    \item We propose the first list-wise ranking paradigm for personality detection, which eliminates the categorical fallacy inherent in existing methods.
    \item We introduce a trait-cohort advantage estimation for the reward model, which calculates the reward score by dimension similarity and NDCG to amplify subtle interactions, generating dense reward signals to effectively avoid the problem of reward hacking. 
    \item Our method achieves state-of-the-art results on mainstream personality detection datasets, establishing a new standard for data-driven personality assessment.
\end{itemize}

\section{Related Work}
\subsection{Personality Detection}
Pre-trained language models (e.g., BERT, RoBERTa) dominated personality detection before the emergence of large language models (LLMs). While demonstrating performance gains through encoding concatenated user utterances, these approaches neglected temporal dynamics and inter-post relationships (Tandera
et al. 2017\nocite{2017Personality}; Xue et al. 2018\nocite{2018Deep}; Lynn, Balasubramanian, and
Schwartz 2020\nocite{lynn-etal-2020-hierarchical}). To address the limitations, Yang et al. (2023a\nocite{yang2023orders}) address the unstructured post-integration challenge in personality detection through Deep Dynamic Graph Convolutional Network (D-DGCN). Departing from prior graph-based methods that require predefined adjacency structures. Concurrently with architectural innovations, LLM-based reasoning techniques emerged, and Chain-of-Thought (CoT) prompting has been introduced to steer LLMs in generating explicit reasoning pathways.   Building upon standard CoT, PsyCoT (Yang et al. 2023b\nocite{yang2023psycot}) innovatively adapts psychological questionnaires into a structured CoT framework. This approach simulates human self-assessment procedures via multi-turn dialogues, directing LLMs to focus on specific test items while leveraging historical ratings for more consistent and accurate reasoning. Hu et al. (2024\nocite{hu2024llm}) address the data scarcity challenge in personality detection through LLM-based text augmentation. Their method distills LLM capabilities (comprehension, summarization, sentiment analysis) to generate multi-perspective text analyses, semantic, affective, and linguistic. Bi et al. (2025\nocite{bi2025llm}) propose the LLM-Enhanced Text Mapping Model (ETM), a hybrid architecture addressing core limitations in traditional personality detection systems. ETM demonstrates how LLMs can dynamically construct psychologically-grounded representations. Although existing methods achieve notable success, they universally adopt classification-centric frameworks, either selecting among 16 discrete MBTI types or aggregating binary predictions across four dimensions. These approaches fundamentally diverge from our work, which harnesses LLMs' reasoning capabilities to reframe personality detection as comparative ranking. By amplifying inter-type distinctions through relative intensity assessments, our paradigm captures nuanced trait interactions obscured by classification methods, thereby advancing detection performance.

\subsection{Fine-Tuning Language Models}
Beyond architectural innovations, training strategies also play a crucial role in enhancing personality detection capabilities (Wang and Sun 2024\nocite{wang2024continuousoutputpersonalitydetection}; Serapio-García et al. 2025\nocite{serapiogarcía2025personalitytraitslargelanguage}). Fine-tuning has emerged as a pivotal strategy for adapting language models to specialized tasks (Shen et al. 2025\nocite{shen2025betterparameterefficientfinetuninglarge}). Recent innovations integrate explicit reasoning traces during fine-tuning, where step-by-step rationales guide models through complex problem-solving (e.g., mathematics or coding) (Wei et al. 2022\nocite{wei2022chain}). These traces mitigate hallucination risks and enhance logical coherence, which proves essential for reasoning scenarios. Reinforcement Learning (RL) further augments reasoning capabilities.       Algorithms like Group Relative Policy Optimization (GRPO) (Guo et al. 2024\nocite{guo2025deepseek}) optimize policies via reward signals. Many works use this training method to fine-tune the domain-specific LLMs. In the financial domain, Liu et al. (2025\nocite{liu2025finr1}) introduced Fin-R1, which employs a two-stage training framework of SFT followed by RL, significantly enhancing its ability to perform complex financial reasoning tasks. Similarly, Papicchio et al. (2025\nocite{papicchio2025think2sql}) proposed Think2SQL, which also utilizes SFT and RL to improve the model's performance in converting natural language questions into SQL queries. These studies collectively demonstrate that the integration of SFT and RL can effectively improve the reasoning capabilities of language models in specific domains, also providing valuable insights for personality detection tasks. Building on these insights for personality detection, our work implements an SFT+GRPO framework with a novel Normalized Discounted Cumulative Gain (NDCG) reward function incorporating MBTI dimensional similarity metrics. This design intrinsically avoids reward hacking through position-aware discounting, enables automatic pattern discovery from behavioral data, resolves expert dependency, and captures cross-dimensional trait interactions, establishing a new paradigm for completed personality-centric reasoning enhancement.

\begin{figure*}[t]
\centering
\includegraphics[width=0.8\textwidth]{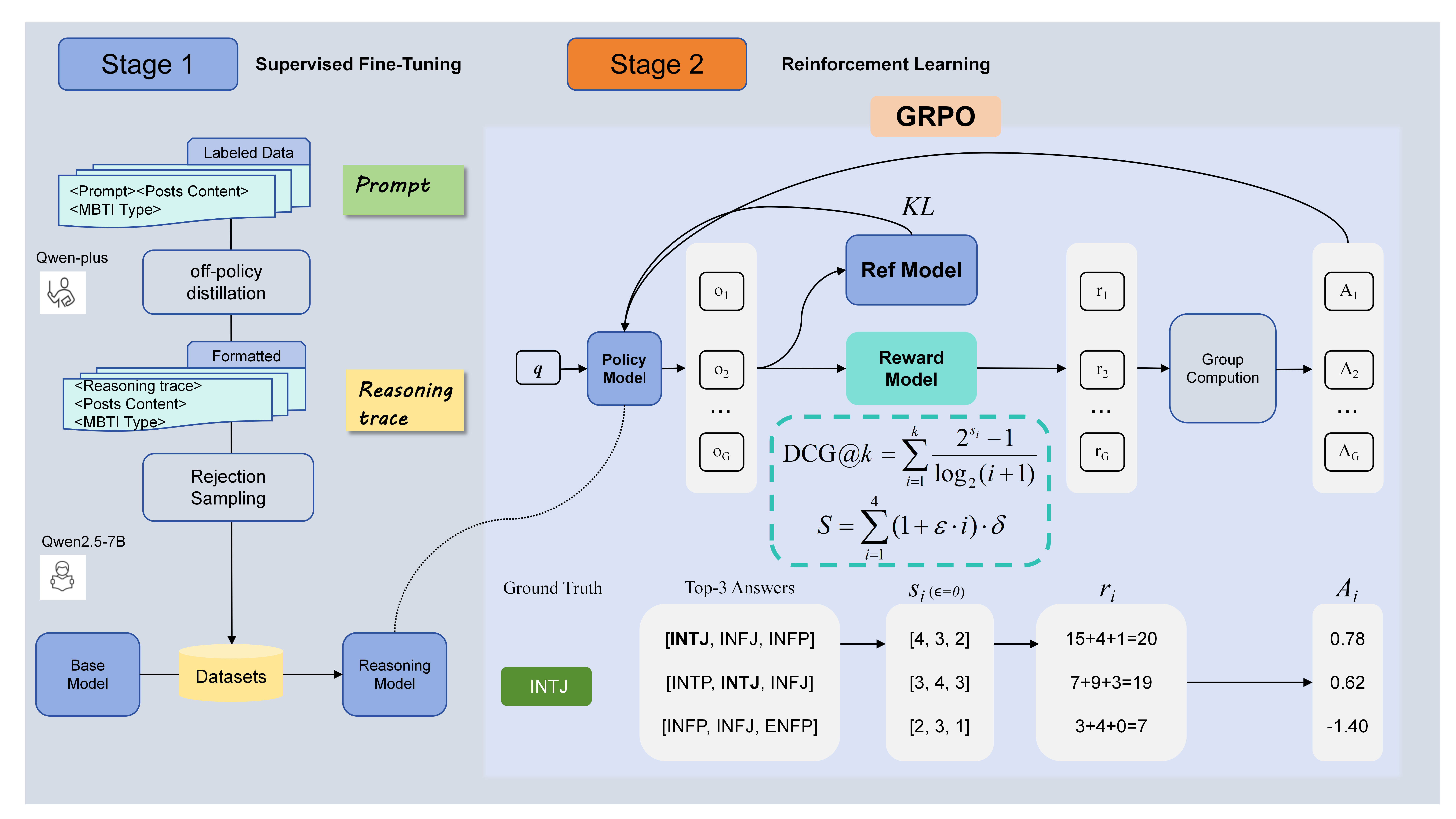} 
\caption{An overview of our PerDet R1, the two-stage LLM post-training workflow. In the first stage, we adopt  Qwen-plus as the teacher model to generate a reasoning trace for comparing different MBTI types. Subsequently, the base model learning formatted data via SFT to get a reasoning model. In the second stage, the reasoning model optimizes via Group Relative Policy Optimization (GRPO) with NDCG and dimension similarity reward function to enhance ranking capability. }
\label{fig2}
\end{figure*}

\subsection{Rule-based Reward Modeling}
Rule-based reward modeling has emerged as a critical paradigm for aligning large language models (LLMs), establishing structured computational frameworks that complement or supplant human preference data to enhance training efficiency and policy controllability (Xiong et al. 2023\nocite{xiong2023iterative}; Tong et al. 2024 \nocite{NEURIPS2024_c4e380fb}; Dong et al. 2024\nocite{dong2024rlhf}).  This approach addresses fundamental limitations identified in the empirical analysis by Gao et al. (Gao, Schulman, and Hilton 2023\nocite{gao2023scaling}) of the overoptimization of the reward model within direct alignment algorithms.  Building on these insights, the Logic-RL framework (Xie et al., 2025\nocite{xie2025logicrl}) pioneers a dual-component reward architecture comprising Format Reward, which enforces structural compliance through formal grammar and reasoning pattern; Validation-Answer Reward, ensuring semantic fidelity via knowledge-constrained verification mechanisms.  This integrated system synergistically guides models toward logically structured reasoning pathways while intrinsically mitigating reward hacking through cross-component dependency checks, where conflicting signals trigger differential reward attenuation.  By leveraging formally specified rules and hierarchically structured rewards, the paradigm establishes robust behavioral guardrails against optimization shortcuts while providing domain-adaptable alignment frameworks and computationally tractable supervision alternatives, demonstrating particular efficacy in domains requiring simultaneous procedural correctness and substantive accuracy.

\section{Method}

We formulate personality detection as a multiclass ranking task, where the goal is to identify and rank the most relevant MBTI types for a user based on their social media posts.  Formally, given a set $P =\{p_1,p_2,...,p_n\} $ of social media posts by a user, the goal is to select the most relevant Top-K MBTI personality option $Y = \{y_1,y_2,...,y_n\}$ for this user based on $P$. As the overall architecture shows in Figure 2, we adopt a two-stage LLM post-training process, namely SFT and Reasoning RL.  Both need to transform the data to a specific format for model fine-tuning.  The first stage focuses on developing LLM’s reasoning abilities and also distills domain knowledge from the teacher model into the student model to achieve the effect of the cold start.  The second stage aims to push LLM to compare the MBTI personality type for the candidate and select the most relevant answers.  Followed by the presentation of some exciting results, and hope this provides the community with valuable insights.

\subsection{Training Data Construction}
The dataset construction involves a rigorous two-stage process. The purpose of each stage of fine-tuning is different, so the form and content of data are also different, which requires the reconstruction and expansion of the dataset to adapt to the fine-tuning requirements of LLM in different stages.

The original dataset $V_{origin}$ is the public dataset that was collected from social media platforms. Each sample $v$ in the original dataset contains two components: the posts and the personality label of this user. This format does not apply to fine-tuning generative language models. To build the dataset $V$ for SFT, we utilize the Qwen-plus model with a specialized prompt, as shown in Figure 3, to generate the reasoning trace and the most relevant answers.  

During the first stage, each sample $v$ in the SFT training datasets $V$ comprises three components, that is, where $P$ denotes the content of the posts formatted as plain text, $R$ represents the reasoning trace, and is formatted as $<$think$>$...$<$/think$>$, $Y$ corresponds to the Top-K answer list, which is formatted as $<$answer$>$[option 1, option 2, ..., option N]$<$answer$>$. Since not all personality types can be accurately inferred from social media posts alone, we apply rejection sampling to filter out samples where the ground truth does not appear in the model's Top-K predictions. By evaluating the model output, the sampling method can filter out low-quality inference chains, retaining only those that meet the standards. This ensures that the model's inference chains are not only correct but also highly readable. Specifically, we filter out the items for which the ground-truth label does not appear in the answer list. Finally, 1,000 samples were carefully selected as high-quality SFT training samples. We use them as initial data for cold start, which helps the model get started without a large amount of supervised data. By using carefully designed small amounts of inference data, the model can produce reasonable inference outputs in the early stages of training.

During the second stage, we use the original datasets that exclude corresponding SFT training samples as the RL training dataset since the cold start data should not appear in this stage. In conclusion, Stage 1 aims to instill basic reasoning patterns from a strong teacher model; Stage 2 uses reinforcement learning to optimize trait inference through reward feedback.

\begin{figure}[t]
\centering
\includegraphics[width=0.9\columnwidth]{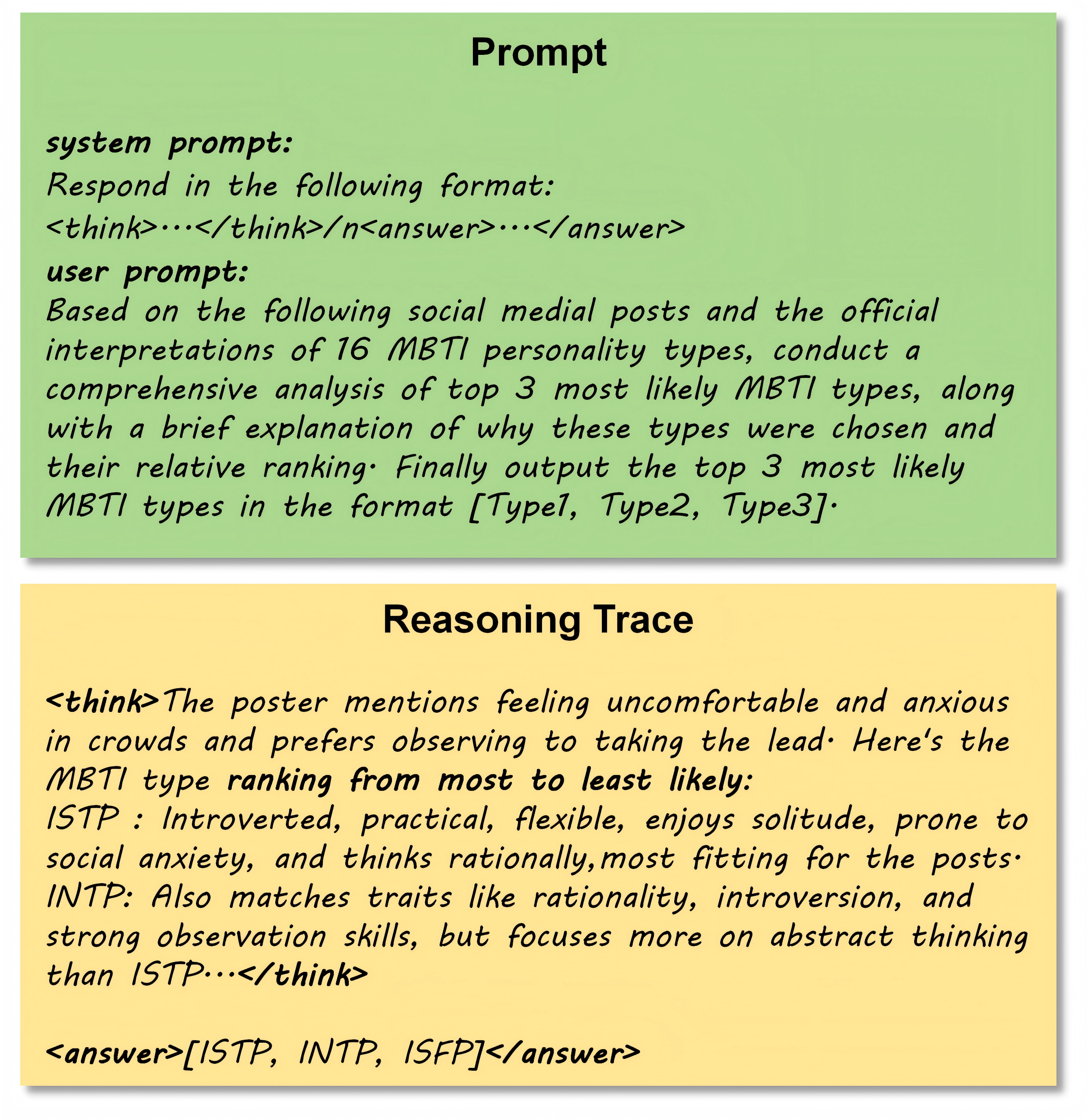}
\caption{An example of prompts and generated reasoning trace with think and answer label for top-k MBTI types ranking. }
\label{fig3}
\end{figure}

\subsection{Supervised  Fine-tuning }
To enhance performance on specific tasks, researchers employ instruction tuning (Chung et al. 2024\nocite{chung2022scaling}), also known as Supervised Fine-Tuning (SFT) (Ouyang et al. 2022\nocite{ouyang2022training}); During this stage, the domain knowledge can be injected to the model (Hsieh et al. 2023\nocite{hsieh2023distilling}) and and enables the model to distinguish different personality traits from social media posts. The objective at this stage is to instill foundational reasoning patterns in the model without overly emphasizing immediate reasoning performance. Additionally, by leveraging the knowledge from large-scale models, computational costs can be reduced. 

We initially performed SFT on the instruct model \texttt{Qwen2.5-7B-Instruct}, and the training data was sampled from the reasoning model \texttt{Qwen-plus}. This method aims to fine-tune the instruct model to replicate the reasoning behavior of the reasoning model, that is, to compare and rank the candidate answers first, and then output the Top-K answers after standardized format processing.

\subsection{Reasoning Reinforcement Learning}
Reinforcement learning (RL) now plays a pivotal role in training LLMs. It is used not only to align responses with human preferences through Reinforcement Learning from Human (RLHF) (Ziegler et al. 2019\nocite{ziegler2020finetuning}; Ouyang et al. 2022\nocite{ouyang2022training}; Lee et al. 2024\nocite{lee2024rlaif}), but also to enhance models’ reasoning abilities, as recently demonstrated by DeepSeek-R1 (Guo et al. 2025\nocite{guo2025deepseek}). These applications underscore RL’s vast potential to drive further advancements in LLMs.

We employ the Group Relative Policy Optimization (GRPO) algorithm, which offers a value-free alternative by computing normalized group-level advantages based directly on realized rewards. Let $\bm{x} \sim \mathcal{X}$ represent the prompt extracted from the conditional input distribution, and $\left\{ \bm{y}_i \right\}_{i=1}^G \sim \pi_{\theta_{\mathrm{old}}}(\cdot \mid \bm{x})$ denote the $G$ response sequences generated by the frozen reference strategy $\pi_{\theta_{\mathrm{old}}}$. For each response $\bm{y}_i = (y_{i,1}, \dots, y_{i,T_i})$, the reward model assigns a scalar reward $R_i \in \mathbb{R}$.
The relative advantage $A_i$ of the $i-th$ response group is defined by normalizing the distribution of group rewards, that is to say, each response receives a reward $r_i$, from which they compute the group-relative advantage $A_i$:
\begin{align}
    A_i = \frac{ r_i - \mu_{r} }{ \sigma_{r} }
    \label{eq:advantage}
\end{align}
where $\mu_{r}$ and $\sigma_{r}$ denote the mean and standard deviation of reward values within the group.

For each token position $t$ in response $y_i$ , define the state as $s_{i,t}=x||y_{i,<t}$ and the token-level probability ratio as:
\begin{align}
    p_{i,t}(\theta) = \frac{
        \pi_{\theta}(y_{i,t} \mid \mathbf{s}_{i,t})
    }{
        \pi_{\theta_{\text{old}}}(y_{i,t} \mid \mathbf{s}_{i,t})
    }
    \label{eq:probability_ratio}
\end{align}
The policy update now maximizes the following objective function:
\begin{align}
    L_{\text{GRPO}}(\theta) = 
    E \Big[ 
        \frac{1}{G} \sum_{i=1}^{G} \frac{1}{T_i} \sum_{t=1}^{T_i}
        \min ( 
            p_{i,t}(\theta) A_i, \ \notag\\
            clip\left( p_{i,t}(\theta), 1-\epsilon, 1+\epsilon \right) A_i 
        ) 
        - \beta KL \bigl[ \pi_{\theta} \parallel & \pi_{\theta_{\text{ref}}} \bigr]
    \Big]
    \label{eq:grpo_loss}
\end{align}
Where $\epsilon$ is the hyper-parameter for clipping parameter and $\beta$ controls the Kullback-Leibler divergence (KL) regularization by penalizing the deviation from the reference policy model.

\subsection{Reward Function Design}

To reduce LLM training costs, GRPO (Shao et al. 2024\nocite{shao2024deepseekmath}) eliminates the value model and relies solely on rule-based reward functions for optimization, making the reward signal critical for guiding exploration.  In MBTI personality detection, a naive approach directly outputting personality types would permit straightforward reward assignment but fundamentally conflicts with the need of the GRPO method for differentiated group-wise rewards.  Classification-based rewards inherently produce sparse and undifferentiated signals, causing minimal score variance within groups and not reflecting relative advantages.  To avoid sparse and undifferentiated reward signals common in classification-based settings, we design a ranking-based reward function using dimension similarity and Normalized Discounted Cumulative Gain (NDCG) that captures both correctness and relevance of the output position.

The NDCG metric is widely adopted in recommendation systems and information retrieval as our reward function. NDCG prioritizes relevance and ranking position, assigning a greater weight to items ranked higher in a list. This effectively measures whether the output list contains the ground-truth and the ordinal position of the correct answer in the output list. During the aforementioned Supervised Fine-Tuning (SFT) phase, the model has learned to generate Top-K MBTI types after reasoning analysis. Subsequently, the MBTI dimension similarity scoring is applied to compute the relevance values, which are then leveraged to calculate the final NDCG-based reward score. The NDCG at $k-th$ position $(k \leq K)$ is defined as:
\begin{align}
\text{NDCG@}k=\frac{\text{DCG@}k}{\text{IDCG@}k}
\end{align}
DCG measures the cumulative benefits of actual ranking and discounts the results that are lower in the ranking:
\begin{align}
\text{DCG@}k=\sum_{i=1}^{k} \frac{2^{s_i} - 1}{\log_2(i + 1)}
\end{align}
IDCG is the DCG value arranged in descending order of correlation, which represents the theoretical optimal result:
\begin{align}
\text {IDCG@}k=\sum_{i=1}^{k} \frac{2^{s^{\text {sorted }}}-1}{\log _{2}(i+1)}
\end{align}
where $s_i$ is the similarity score for ground-truth and the i-th predicted type in the Top K list. Since MBTI consists of four dimensions, each of dimension contains two opposing endpoints, resulting in sixteen different personality types. For these characteristics, we calculate similarity scores using the following formula:
\begin{align}
\mathrm{s}\left(\hat{\mathrm{t}}, \mathrm{t}^{*}\right)=\sum_{i=1}^{4}(1+\epsilon \cdot i) \cdot \delta\left(\hat{t}^{(i)}, t^{*(i)}\right)
\end{align}
Where $\epsilon \in \mathbb{R}^+$ denotes the dimension weight coefficient, a hyper-parameter controlling the contribution strength of each dimension’s character match, let $T = \{\hat{t}_1, \hat{t}_2, \dots, \hat{t}_k\}$ denotes the list of predicted MBTI types containing $k$ elements, where each $\hat{t}_j$ represents a complete personality type prediction with four-letter format. And $t^*$ represents the ground-truth from datasets. $\delta$ is the character matching function, defined as:
\begin{align}
\delta(a,b) = \left\{
\begin{array}{ll}
1 & , a = b \\
0 & , a \neq b 
\end{array}
\right.
\end{align}
where $\delta$ measures the degree of match between the predicted and the ground truth.
To avoid the problem of reward hacking in rule-based reward model, We also use the $\mathrm{s}\left(\hat{\mathrm{t}_1}, \mathrm{t}^{*}\right)$ as our Dimension Similarity Reward (DS Reward) which denotes dimension similarity of ground-truth and the first answer from top k predict list to enhance the effect of the first answer.We take the sum of NDCG and DS as the total reward function.

\begin{table}[htbp]
\centering
\resizebox{0.48\textwidth}{!}{
\label{tab:results}
\begin{tabular}{lrrrr}
\toprule
\multirow{2}{*}{\centering \textbf{Methods}} & \multicolumn{2}{c}{\textbf{Kaggle}} & \multicolumn{2}{c}{\textbf{PANDORA}} \\
\cmidrule(lr){2-3} \cmidrule(lr){4-5}
& \begin{tabular}{@{}c@{}}Macro-F1\\(Binary)\end{tabular} & \begin{tabular}{@{}c@{}}F1-score\\(Multi)\end{tabular} & \begin{tabular}{@{}c@{}}Macro-F1\\(Binary)\end{tabular} & \begin{tabular}{@{}c@{}}F1-score\\(Multi)\end{tabular} \\
\midrule
SVM & 60.21 & 22.39 & 53.15 & 18.55 \\
BERTconcat & 60.61 & 23.77 & 53.91 & 10.03 \\
BERTmean & 66.04 & 26.38 & 56.52 & 17.56 \\
D-DGCN & 71.35 & 30.32 & 61.40 & 29.69 \\
ChatGPT & 63.89 & 34.55 & 60.08 & 29.87 \\
Qwen-plus & 76.43 & 38.63 & 61.34 & 25.08 \\
PsyCoT & 65.22 & 31.22 & 60.81 & 29.81 \\
TAE & 72.07 & 32.09 & 63.05 & 30.22 \\
ETM & 77.79 & 32.55 & 65.77 & 30.09 \\
\textbf{PerDet-R1 (our)} & \textbf{80.57} & \textbf{41.34} & \textbf{66.10} & \textbf{35.08} \\
\bottomrule
\end{tabular}
}
\caption{Performance comparison on Kaggle and Pandora datasets. The average Macro-F1 for the 4-dimensional binary classification task and F1-Score for 16-types multiclass classification task.}
\end{table}

\section{Experiments}
\subsection{Datasets}
Following previous studies (Yang et al. 2023a; Hu et al. 2024; Bi et al. 2025), we conducted experiments on the most commonly used personality detection datasets, including Kaggle\footnote{https://www.kaggle.com/datasets/datasnaek/mbti-type} and PANDORA\footnote{https://psy.takelab.fer.hr/datasets/all/}. These datasets cover mainstream social platforms and forums. The Kaggle dataset was collected through the PersonalityCafe\footnote{http://personalitycafe.com/forum} forum, which is focused on personality type discussion. It comprises a total of  8675 users’ data, each consisting of the 50  most recent posts. PANDORA was collected from Reddit\footnote{https://www.reddit.com/}, with dozens to hundreds of social media posts for each of 9067 users. As in previous work (Yang et al. 2023a\nocite{yang2023orders}), we remove words that match any personality type from the posts to avoid label leakage. Performance is evaluated using the average Macro-F1 metric, which is the same as in previous work. We take the first answer in the top 3 MBTI answer list as the predicted value to calculate the average Macro-F1 for the 4-dimensional binary classification task, F1-Score for the 16 MBTI types multiclass classification task, and NDCG@$k$ metric for the ranking task. We opted for Macro-F1 to ensure fairer comparisons with prior research findings. Furthermore, since our approach treats the four MBTI dimensions as a unified label, we employ F1-Score to comprehensively evaluate performance in multi-class classification tasks. Meanwhile, applying NDCG not only effectively reflects the training outcomes of stage 2 but also demonstrates whether the ranking task successfully ranks answers closest to the ground truth in the top position.

\subsection{Baselines}
To evaluate the proposed models intensively, we employ the following mainstream models as our baseline. 

\noindent \textbf{SVM} (Cui and Qi 2017\nocite{cui2017survey}) and XGBoost (Tadesse et al. 2018\nocite{tadesse2018personality}): These methods concatenate all the posts of a user 
into a document first, and then employ SVM or XGBoost 
for classification based on features extracted using bag-of-words
word models. They represent the advanced performance of traditional machine learning algorithms in personality detection tasks.

\noindent \textbf{BERT-concat} (Jiang, Zhang, and Choi 2019\nocite{jiang2020automatic}): This method concatenates a user’s posts into an extended document and then utilizes BERT to encode this composite text for user representation. 

\noindent \textbf{BERT-mean} (Keh and Cheng 2019\nocite{keh2019myers}): It uses BERT to encode each post individually, applies average pooling to get user feature embedding, and then maps the features to personality labels.

\noindent \textbf{D-DGCN} (Yang et al. 2023a\nocite{yang2023orders}): D-DGCN employs a dynamic graph to encode each post via domain-adapted BERT, then employs DGCN to encode the graph and obtains a user representation. And it’s the SOTA method (Kaggle dataset) in a BERT-based model.

\noindent \textbf{PsyCoT} (Yang et al. 2023b\nocite{yang2023psycot}): This method prompts the LLM to engage in reasonable personality reasoning. Introducing items from the questionnaire as a set of rigorous CoT. It guides the LLM to evaluate each item based on the author’s text and then infer the final personality trait.

\noindent \textbf{ChatGPT}: We applied the ’gpt-3.5-turbo-0301’ ver-sion 
of ChatGPT, and set the temperature to 0, making the outputs mostly deterministic for the identical inputs.

\noindent \textbf{Qwen-plus} (Yang et al. 2025\nocite{yang2025qwen3}): We applied the Qwen3 flagship model, and set the temperature to 1, making the outputs mostly deterministic for the identical inputs.

\noindent \textbf{TAE} (Hu et al. 2024\nocite{hu2024llm}): This method distills the useful knowledge from the LLM to address the data scarcity issue faced by small models in personality detection. It utilizes LLM to generate posts analysis from semantic, sentiment, and linguistic aspects, enhancing data and personality label representations.

\noindent \textbf{ETM} (Bi et al. 2025\nocite{bi2025llm}): This method leverages the text embedding and text generation capabilities of LLM to address the issues of poor user vector representation and the insufficient relationship between user vectors and personality labels in small model-based personality detection.

\subsection{Implementation Details}
For the traditional deep learning method, we use PyTorch and use Adam Optimizer (Kingma and Ba 2015\nocite{kingma2015adam}), the same as in the previous study. For LLM, we set up a batch size of 256 in the SFT stage and 128 in the RL stage with several generations of 16. The temperature parameter ($\tau$) is maintained at 1. The fine-tuning process was conducted under the LoRA framework (Hu et al. 2021\nocite{hu2022lora}). We selected a rank of 32 and conducted the SFT throughout 3 epochs and RL over 2000 steps. The top k parameter is set to 3, and the dimension weight coefficient parameter $\epsilon$ is set to 0.1. Each post is limited to 128 tokens. We set the maximum number of posts per user to 50. We replace words that match any personality label with the $<$MASK$>$ to avoid information leaks (Yang et al. 2023a\nocite{yang2023orders}). 
We use \texttt{Qwen-plus} for distilling reasoning trace and \texttt{Qwen2.5-7B-Instruct} as the base model that needs to be fine-tuned.

\begin{table}[ht]
\centering
\resizebox{0.48\textwidth}{!}{
\label{tab:results}
\begin{tabular}{@{}lc ccc@{}}
\toprule
\multirow{2}{*}{\centering \textbf{Model} }  &
\multirow{2}{*}{\centering \textbf{Strategy} } 
    & \multicolumn{3}{c}{Kaggle}  \\
    \cmidrule(lr){3-5}
   & & \textbf{Binary} & \textbf{Multi-class} & \textbf{NDCG@3} \\
\midrule
\multirow{3}{*}{\centering Llama3-8B} 
  & w/o SFT, GRPO & 35.29 & 8.80 & 51.15 \\
  & w/o GRPO      & 59.65 & 13.77 & 62.69 \\
  & SFT+GRPO      & 72.57 & 37.88 & 71.64 \\
\midrule
\multirow{8}{*}{\centering Qwen2.5-7B}
  & w/o SFT, GRPO & 35.67 & 6.21 & 49.46 \\
  & w/o SFT       & 68.11 & 17.21 & 67.58 \\
  & w/o GRPO      & 65.03 & 15.82 & 64.09 \\
  & classification & 63.07 & 33.67 & -- \\
  & w/o DS reward & 70.06 & 28.64 & 69.38 \\
  & w/o NDCG reward & 68.90 & 27.28 & 66.92 \\
  & Top-5         & 76.44 & 43.58 & 74.91 \\
  & SFT+GRPO      & 80.57 & 41.34 & 74.31 \\
\bottomrule
\end{tabular}
}
\caption{Results of ablation study over different base models and training strategies, we mainly carry out ablation experiments on Qwen model and Kaggle dataset}
\end{table}

\begin{figure}[t]
\centering
\includegraphics[width=0.9\columnwidth]{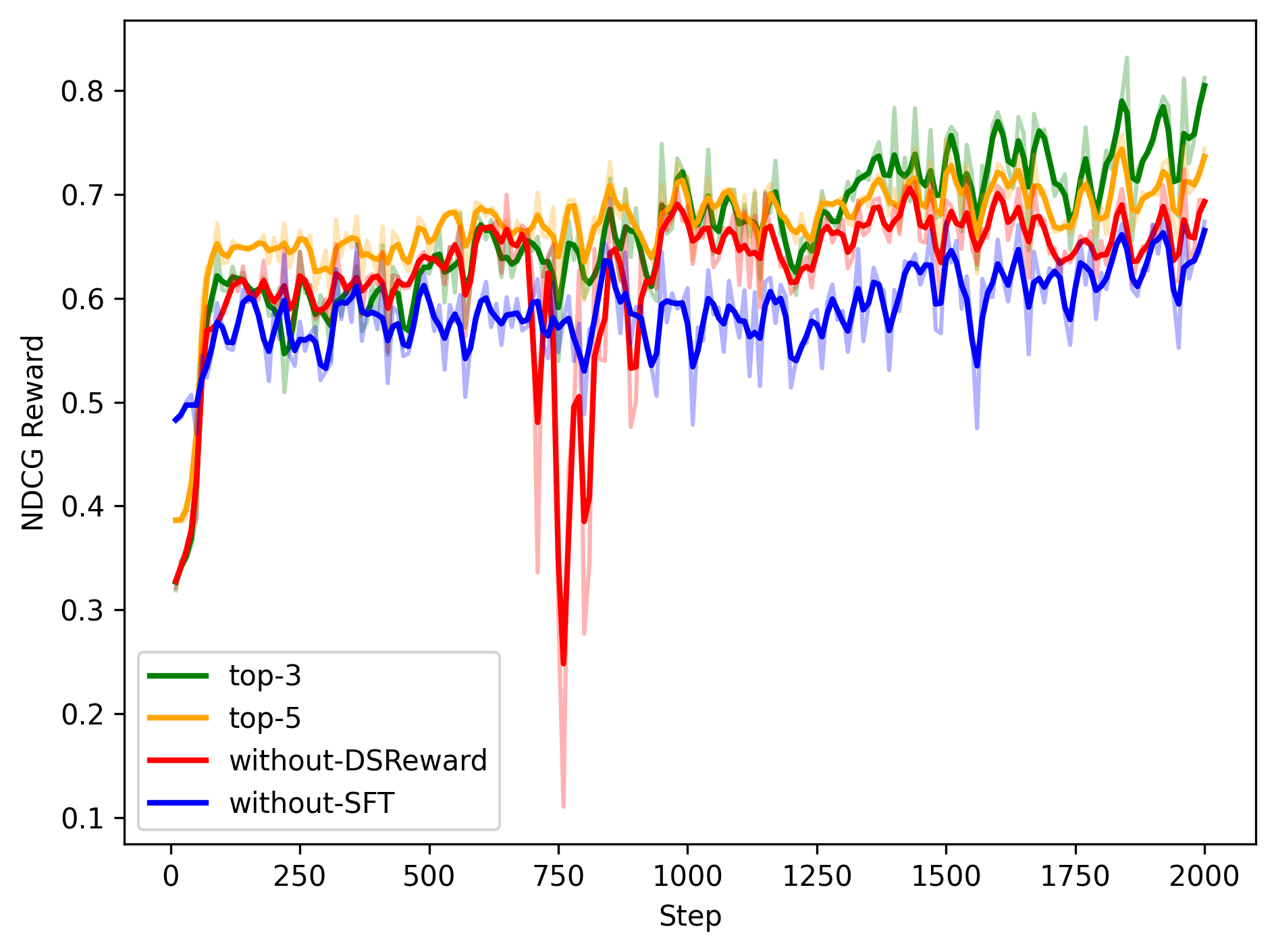}
\caption{NDCG Reward (smoothed with 0.8 sigma) comparison during RL training stage on Kaggle Dataset. }
\label{fig4}
\end{figure}

\subsection{Overall Results}
Table 1 demonstrates that PerDet-R1 outperforms all baselines in binary Macro-F1 on benchmark datasets. Specifically, on the Kaggle and PANDORA datasets, compared with the currently best-performing ETM, our method achieved improvements of 2.78\% and 0.83\% in Macro-F1 scores, as well as 8.79\% and 4.99\% in F1-score scores. Although our primary objective is to rank complete MBTI types in the top position, this design simultaneously enhances the classification capability. It is worth noting that when evaluating personality trait detection as a multi-classification task, PerDet-R1 achieves a significantly higher F1-score compared to other methods. This advantage stems from modeling all four dimensions of MBTI as an integrated element during both two-stage training stages, along with full comparisons of differences between distinct categories.  Overall, our method achieves outstanding performance due to two key enhancements. Firstly, during the SFT stage, it memorizes the strategy and thought process of analysis and comparison through rejection sampling and off-policy distillation, thereby acquiring reasoning capabilities. In the RL stage, it autonomously explores effective paths via the GRPO method, enhancing the model's generalization ability. Secondly, the NDCG-based reward function ingeniously integrates the two stages, modeling personality detection as a ranking task, thereby overcoming the fuzzy boundary issue caused by existing methods that simplistically separate the four dimensions into multiple binary classification tasks.

Through comparative analysis of two datasets, we observed that existing methods generally demonstrate lower accuracy on the Pandora dataset compared to the Kaggle dataset. This discrepancy likely stems from the fact that the Kaggle dataset originates from the MBTI forum, where users frequently employ abbreviations like "NS", "ExFJ", and "xNTP" during discussions. While these terms are not exhaustive answers, they effectively provide strong contextual cues that enhance the data-driven method.

\subsection{Ablation Study}
Table 2 presents the ablation results on the Kaggle dataset to facilitate a comprehensive investigation. PerDet-R1 integrates a two-stage training process (SFT followed by RL), a specially adapted reward function to enhance LLM reasoning capabilities, and a ranking-based inference strategy that replaces classification. PerDet-R1 achieves comprehensive performance improvement through integrated strategies, where removing any component causes overall performance degradation.

\noindent \textbf{Training Strategy}. For training stages, after removing GRPO, the NDCG score decreased by 
8.95\% for Llama3-8B and 10.22\% for Qwen2.5-7B. When removing both SFT and GRPO, the NDCG score dropped substantially by 20.49\% and 24.85\%, respectively, on both models. As shown by the blue line in Figure 4, without the first-stage SFT, the performance improvement during later RL training becomes weak. This demonstrates the necessity of our two-stage training. Meanwhile, we conducted experiments using classification with SFT+GRPO instead of ranking on Qwen2.5-7B, which resulted in decreases of 17.50\% in binary F1 and 7.67\% in multi-class Macro-F1, confirming the significance of the ranking method.

\noindent \textbf{Reward model}. For the reward model in GRPO, we found that removing the dimension similarity reward let to NDCG reward decreased by 4.93\% and 7.39\%, respectively. Furthermore, as observed from the red line in Figure 4, the NDCG reward collapses abruptly during training without the dimension similarity reward. Similarly, the dimension similarity reward collapses when the NDCG reward is removed. This phenomenon confirms that both reward functions are essential for stable performance improvement.

\noindent \textbf{Top-k}. We used a variant number top k with k = 5 to verify whether a larger answer space facilitates exploration. The results show that while binary F1 decreased by 4.13\%, multi-class Macro-F1 and NDCG increased by 2.24\% and 0.60\%, respectively.

These results demonstrate that our method leverages two-stage training and specially adapted reward modeling to enhance LLMs' reasoning capabilities in personality detection tasks significantly.

\section{Conclusion}
In this paper, we propose a list-wise ranking fine-tuning framework for personality detection that establishes a paradigm shift from classification on dimensions to personality trait ranking.  Our method significantly enhanced LLM reasoning capabilities, particularly valuable when models inherently struggle with the task, by fundamentally reconceptualizing personality detection as a comparative ranking problem rather than disjoint binary classifications.  This paradigm shift overcomes the critical limitation of existing approaches, which artificially fragment MBTI dimensions and disrupt their intrinsic interrelationships.  Through a two-stage architecture featuring: 1) SFT with off-policy distillation for reasoning trace generation and rejection sampling for quality control, followed by 2) GRPO optimization with an NDCG-based reward function for generalization enhancement, our approach comprehensively models personality trait inter-class differences.  Experimental results across benchmark datasets demonstrate state-of-the-art performance, validating the framework's efficacy. In future work, we will explore how to achieve higher-performance personality trait detection by utilizing multi-modal data such as profile avatars and facial expressions.

\section{Acknowledgments}
Author Y. Zhu gratefully acknowledge the financial support from the National Key R\&D Program of China (Grant No. 2023YFC2506600, 2023YFC2506601).

\bibliography{aaai2026}


\end{document}